# DICTIONARY LEARNING BASED IMAGE ENHANCEMENT FOR RARITY DETECTION

*Hui Li, Xiaomeng Wang, Weifeng Liu and Yanjiang Wang*

***ABSTRACT***

Image enhancement is an important image processing technique that processes images suitably for a specific application e.g. image editing. The conventional solutions of image enhancement are grouped into two categories which are spatial domain processing method and transform domain processing method such as contrast manipulation, histogram equalization, homomorphic filtering. This paper proposes a new image enhance method based on dictionary learning. Particularly, the proposed method adjusts the image by manipulating the rarity of dictionary atoms. Firstly, learn the dictionary through sparse coding algorithms on divided sub-image blocks. Secondly, compute the rarity of dictionary atoms on statistics of the corresponding sparse coefficients. Thirdly, adjust the rarity according to specific application and form a new dictionary. Finally, reconstruct the image using the updated dictionary and sparse coefficients. Compared with the traditional techniques, the proposed method enhances image based on the image content not on distribution of pixel grey value or frequency. The advantages of the proposed method lie in that it is in better correspondence with the response of the human visual system and more suitable for salient objects extraction. The experimental results demonstrate the effectiveness of the proposed image enhance method.

## 1. INTRODUCTION

Image enhancement plays a fundamental role for image/video processing. The purpose of image enhancement is to process an image so that the result is more suitable than the original image for a specific application such as image editing, medical image processing, remote sensing image processing, and super-resolution reconstruction. Briefly, the conventional image enhance methods are classified into two groups: (1) spatial domain method and (2) transform domain method [1].The former deals with image pixel gray directly, for instance, contrast stretching [2], histogram equalization [3] and other kinds of improved algorithm based on it. Tarik Arici [4] presented a general framework based on histogram equalization for image contrast enhancement.  The latter is mainly based on the Fourier transform of the image to improve the image spectrum by enhancing or inhibiting part of the spectrum, such as low-pass filtering technology, high-pass filtering technology and homomorphic filtering[5].The typical transform domain method is frequency filtering [6]. Muhammad Zafar Iqbal [6] combined dual-tree complex wavelet transform with nonlocal means for resolution enhancement of satellite images.

However, most of the current image enhance methods convert image based on the distribution of pixel grey value or frequency, which may reduce or remove some information concerned. For example, a high-pass filtering operation may weaken objects that contain low frequency information. Although there are some local enhancement methods (e.g. local contrast enhancement [7]), the information of interested objects cannot be well highlighted.

In this paper, a new image enhance method is proposed to well boost the image saliency based on dictionary learning. In particular, the dictionary is learned from the sub-image blocks. The dictionary implies direct relevance to the image content. Therefore, the rarity of the image content can be manipulated by adjusting the dictionary atoms. The advantages of the proposed image enhancement lie in (1) it directly reveals the relation between saliency information and dictionary atoms that is in better correspondence with the response of the human visual system ; (2) it well boosts the saliency and hence it is very suitable for some practical applications such as salient object extraction. Some image enhance experiments are conducted and the results show the effectiveness of the proposed method.

The remainder of the paper is organized as follows. In section 2, the sparse coding algorithm is briefly introduced. Section 3 describes the proposed algorithm in detail. The corresponding experimental results are shown in Section 4. Finally, Section 5 draws the conclusion.

## 2. THE SPARSE CODING THEORY

In the field of computer vision, there has been an increasing interest in the research of sparse representation [8] of signals. The basic idea of the sparse representation is the natural signal could be compressed or could be represented by the linear combination of predefined atoms. Taking an over-complete dictionary matrix $D \in R^{n \times K}$ that contains $K$ atoms, it is supposed that the signal $Y \in R^n$ can be represented as a sparse linear combination of these atoms [9]. The representation of $Y$ could be labeled as (1):

$$Y = Dx \quad \text{or} \quad Y \approx Dx. \tag{1}$$

The vector $x \in R^K$ denotes the representation coefficients of the signal $Y$. Constraints on the solution must be set, whereas an infinite number of solutions are available if $N < K$ and $D$ is a full-rank matrix. The solution with the fewest number of nonzero coefficients is certainly an appealing representation. The sparsest representation is the solution of:

$$\min_x \|x\|_0 \quad s.t. \|Y - Dx\|_2 \leq \varepsilon.. \tag{2}$$

There are many algorithms about sparse coding on the basis of over-complete dictionary, including matching pursuit (MP) algorithm [10], basic pursuit (BP) algorithm [11], and orthogonal matching pursuit (OMP) algorithm [12] etc. Among them, the OMP algorithm chooses the atoms from the over-complete dictionary which have the maximum residual orthogonal projection with the input signal, and handles the atoms by Schmidt orthogonalization, it guarantees the optimality of the iterations, and has smaller number of iterations.

## 3. THE IMAGE ENHANCE METHOD BASED ON DICTIONARY LEARNING

Here we present a new image enhance method based on dictionary learning, which adjusts the image by manipulating the rarity of dictionary. The method is in better correspondence with the response of the human visual system and more suitable for salient objects extraction. The mainly parts are as follows:

### 3.1. Dictionary learning

In the proposed method, the input image is first divided into sub-image blocks (as shown in Fig. 1) and then a sparse code for each block and a corresponding dictionary are learned by sparse coding theory[13]-[14].

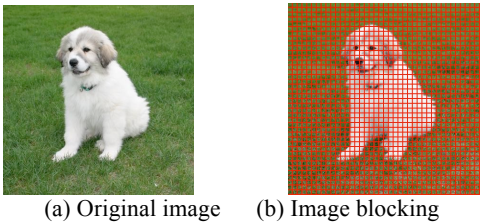

(a) Original image    (b) Image blocking

**Fig. 1.** Image Blocking.

Suppose matrix $Y = \{y_1, y_2, \cdots, y_N\} \in R^{n \times N}$ stands for the $N$ sub-image blocks, matrix $X \in R^{K \times N}$ denotes the sparse codes matrix corresponding to the dictionary $D \in R^{n \times K}$, where $y_i \in R^{n \times 1} (i = 1, 2, \cdots N)$ is the $i^{th}$ sub-block and $K$ is the number of dictionary atoms. Then the sparse codes $X$ and dictionary $D$ can be obtained by solving the following optimization problem:

$$\min_{X,D} \|Y - DX\|_F^2 + \alpha \|X\|_1, s.t. \|D_i\|_2 \leq 1, i = 1, 2, \cdots, K \tag{4}$$

Equation (4) can be easily solved by combining alternating optimization and $l_1$ norm regularization optimization algorithms. Briefly, the optimization algorithm is divided into two alternated optimization sub-problems: sparse coding stage i.e. optimizing with fixed and dictionary update stage i.e. optimizing with fixed. The number of optimization algorithms is numerous, such as the maximum likelihood method [15], [16], method of optimal directions (MOD) [17] and K-SVD [9] are more popular. Here, the K-SVD algorithm is used in this paper. It is flexible, efficient and works in conjunction with any pursuit algorithms, for instance, the MP and the OMP algorithm mentioned above.

### 3.2. Rarity manipulation

The rarity of dictionary atoms can be evaluated from sparse codes. Here, we treat the frequency of one dictionary atom appearing in sub-image blocks as the measurement of its rarity. Higher frequency means less rarity i.e. if an atom appears more often in sub-image blocks, the features it represents are less rare.

Define $R = [R_1 \quad R_2 \quad \cdots \quad R_K]^T$ as rarity measurement of the dictionary atoms where $R_i$ represents the rarity of the $i^{th}$ dictionary atom. There are many ways to compute $R$ from $X$. Suppose $m_i$ is the number of non-zero coefficients in the $i^{th}$ row of $X$. The rarity of the $i^{th}$ dictionary atom can be computed as:

$$R_i = \frac{m_i}{s}, \tag{5}$$

or

$$R_i = \frac{\sum_{j=1}^{N} X_{ij}}{s}. \tag{6}$$

Where, $s$ is a scale constant. It should note that $R_i$ can be expressed in other forms e.g. $R_i = -\log\left(\frac{m_i}{s}\right)$, $R_i = \left(\frac{m_i}{s}\right)^2$.

Then define a rarity transform function $f: R \mapsto \widetilde{R}$ that adjusts the rarity of image content. Then the dictionary can be updated using $\widetilde{D} = D diag(\widetilde{R})$, where $diag(\widetilde{R})$ is a diagonal matrix with $\widetilde{R}_i$ as diagonal entries. And the reconstructed image by $\widetilde{Y} = \widetilde{D}X$ will be enhanced with image rarity changed. There are many ways to manipulate the rarity. One representative form of $f$ is sigmoid function. Carefully designed transform function will be very suitable for specific applications.

### 3.3. Algorithm

In this section, we summarize the algorithm of the proposed image enhancement as follows:

Step 1: Divide the input image into $8 \times 8$ sub-image blocks and form matrix $Y$. Initialize the number of dictionary atoms $K$, e.g. $K$ is set to 256 in the experiments of this letter.

Step 2: Compute sparse codes $X$ and the corresponding dictionary $D$ by solving problem (4) using proper optimization method, e.g. K-SVD is employed here.

Step 3: Compute rarity measurement $R$ based on the statistical analysis of sparse codes matrix $X$.

Step 4: Manipulate rarity $R$ to $\widetilde{R} = f(R)$ according specific application.

Step 5: Update dictionary using $\widetilde{D} = D diag(\widetilde{R})$.

Step 6: Calculate the result image using $\widetilde{Y} = \widetilde{D}X$.

### 4. EXPERIMENTS AND DISCUSSIONS

In order to verify the proposed method, experiments are carried out under 64-bit Windows 7 system. The computer is configured with Intel 2.5 GHz processor, 2GB RAM, MATLAB R2012a software platform. The test images used are from the MSRA salient object database[1]. Experiments of image rarity detection are conducted using different sigmoid transform functions.

Fig.2 shows the example of the proposed image enhancement. In Fig.2, a shows the original image, b displays the dictionary of the image. while c and d are the enhanced image results. From the experimental results, we can see that the proposed image enhancement can adjust images according to the image saliency.

---

[1] http://research.microsoft.com/en-us/um/people/jiansun/SalientObject/salient_object.htm

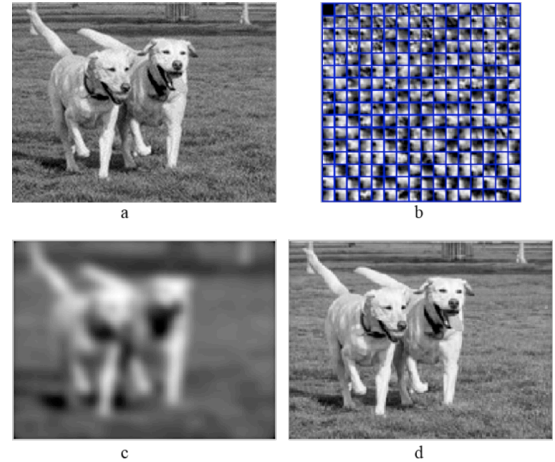

**Fig. 2.** An example of the proposed method.

Image rarity detection experiments are also conducted comparing with the representative Itti [18] and GBVS [19] model. Fig.3 shows some experimental results. Subfigures of row 1 have static backgrounds. Subfigures of row 2 have dynamic backgrounds e.g. water waves, swing grasses. Subfigures of row 3 contain multiple salient objects. In Fig.3, Column a and e are original images, Column b and f are the results of ITTI method, Column c and g are the results of GBVS method, Column d and h are the results of our method. From the comparison, we can see that our method can get the more accurate shape and size of the salient objects. In particular, the results of our methods are comparable with ITTI and GBVS methods when there is static background. The results of our methods are better when there is dynamic background. Furthermore, our method can better discriminate multiple objects than other methods.

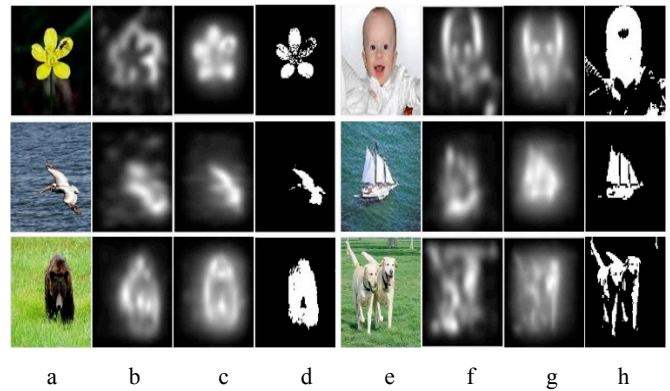

**Fig. 3.** Experimental results.

### 5. CONCLUSION

Here we propose a new image enhance method based on dictionary learning. By adjusting the rarity of dictionary atoms, the proposed method can enhance images directly according image content. Experimental results show that the proposed image enhancement can well steer the image saliency and is very suitable for objects of interest extraction.

**REFERENCES**

[1] N. P. Galatsanos, C. A. Segall, and A. K. Katsaggelos, "Digital image enhancement," *Encyclopedia of Optical Engineering*, pp. 388-402, 2003.

[2] A. Toet, "Multiscale contrast enhancement with applications to image fusion," *Optical Engineering*, vol.31, pp.1026-1031, 1992.

[3] S.M. Pizer, E.P. Amburn, J.D. Austin, R. Cromartie, A. Geselowitz, T. Greer, B.H. Romeny, J.B. Zimmerman, and K. Zuiderveld, "Adaptive histogram equalization and its variations," *Computer Vision, Graphic, and Image Processing*, vol.39, pp. 355–368,1987.

[4] T. Arici, S. Dikbas, and Y. Altunbasak, "A histogram modification framework and its application for image contrast enhancement," *IEEE Transactions on image processing*, vol. 18, no. 9, pp. 1921-1935, Sept. 2009 .

[5] R. W. Fries and J.W. Modestino, "Image enhancement by stochastic homomorphic filtering", *IEEE Transactions on Acoustics, Speech and Signal Processing*, vol. 27, pp. 625-637,1979.

[6] M. Z. Iqbal, A. Ghafoor, and A. M. Siddiqui, 'Satellite image resolution enhancement using dual-tree complex wavelet transform and nonlocal means', *IEEE Geoscience and Remote Sensing Letters*, vol.10, pp. 451-455, 2013

[7] C. Deng, X.-B Gao, et al. "Local histogram based geometric invariant image watermarking," *Signal Processing*, Vol.90, No.12, pp.3256-3264, 2010.

[8] J. Yu, Y. Rui, and D. Tao, "Click prediction for web image reranking using multimodal sparse coding," *IEEE Transactions on Image Process*, vol. 23, pp.2019-32, 2014.

[9] M. Aharon, M. Elad, and A Bruckstein, "K-SVD: An algorithm for designing overcomplete dictionaries for sparse representation," *IEEE Transactions on Signal Processing*, vol. 54, pp.4311-4322, 2006.

[10] S. Mallat and Z. Zhang, "Matching pursuits with time-frequency dictionaries," *IEEE Transactions on Signal Processing*, vol. 41, pp.3397-3415, 1993.

[11] S.chen, D. Donoho, and M. Saunders, "Atomic decomposition by basis pursuit," *SIAM Review*, vol. 43, no. 1, pp. 129–159, 2001.

[12] Y. C. Pati, R. Rezaiifar, and P. S. Krishnaprasad, "Orthogonal matching pursuit: Recursive function approximation with applications to wavelet decomposition," *Proceedings of the 27th Asilomar Conference on Signals, System and Computers*, pp.40-44, 1993.

[13] W. F. Liu, D. C. Tao, J. Cheng, and Y. Y. Tang, "Multiview hessian discriminative sparse coding for image annotation," *Computer Vision and Image Understanding*, vol.118, pp.50-60, 2014.

[14] W. F. Liu, H. M. Zhang, D. P. Tao, Y. J. Wang, and K. Lu, "Large-scale paralleled sparse principal component analysis," *Multimedia Tools and Applications*, 1-13, 2013.

[15] B. A. Olshausen and D. J. Field, "Sparse coding with an overcomplete basis set: A strategy employed by V1?," *Vision Research*, vol. 37, pp. 3311-3325, 1997.

[16] M. S. Lewicki and B. A. Olshausen, "Probabilistic framework for the adaptation and comparison of image codes," *Journal of the Optical Society of America A* , vol. 16, pp. 1587-1601, 1999.

[17] K. Engan, S. Aase, and J. Husoy, "Method of optimal directions for frame design," *Proceedings of IEEE International Conference on Acoustics, Speech, and Signal Processing*, 1999, vol.5, pp. 2443-2446.

[18] L. Itti, C. Koch, and E. Niebur, "A model of saliency-based visual attention for rapid scene analysis," *IEEE Transactions on pattern analysis and machine intelligence*, vol. 20, no. 11, pp. 1254-1259, 1998.

[19] J. Harel, C. Koch and P. Perona, "Graph-based visual saliency," *Proceedings of neural information processing systems(NIPS)*, 2006.